\documentclass[letterpaper, 10 pt, conference]{./support/ieeeconf}
\usepackage{times}
\usepackage[pdftex]{graphicx}
\usepackage{subfigure}
\usepackage{amsmath,amssymb,amsopn,amstext,amsfonts}
\usepackage{cancel}
\usepackage[space]{cite}
\usepackage{pdfsync}
\usepackage{balance}
\usepackage{color}
\usepackage{mathtools}
\usepackage[ruled,vlined]{algorithm2e}
\usepackage{bm}

\usepackage{diagbox}
\usepackage{float}
\usepackage{epstopdf}
\usepackage{pifont}
\usepackage{fixltx2e}
\usepackage{amsmath}
\usepackage{multirow}
\usepackage{booktabs}
\usepackage{url}
\usepackage{svg}
\usepackage{threeparttable}
\usepackage[linkcolor=black,citecolor=black,urlcolor=black,colorlinks=true]{hyperref}
\newcommand{\tp}{^{\mathrm{T}}}

\newcommand{\df}[1]{\mathrm{d}{#1}}
\newcommand{\rbrac}[1]{({#1})}

\newcommand{\cBrac}[1]{\left\{{#1}\right\}}
\newcommand{\sbrac}[1]{[{#1}]}

\newcommand{\norm}[1]{\Vert{#1}\Vert}

\graphicspath{{./figures/}}
\DeclareGraphicsExtensions{.png,.jpg,.eps,.pdf}
\IEEEoverridecommandlockouts
\makeatletter
\def\endthebibliography{%
  \def\@noitemerr{\@latex@warning{Empty `thebibliography' environment}}%
  \endlist
}

\title{\LARGE \bf Elastic Tracker: A Spatio-temporal Trajectory Planner\\ for Flexible Aerial Tracking}
\author{Jialin Ji, Neng Pan, Chao Xu, and Fei Gao
\thanks{Corresponding Author: Fei Gao,  \tt\small{fgaoaa@zju.edu.cn}}
\thanks{This work was supported by the National Natural Science Foundation of China under Grants 62003299.}
\thanks{The State Key Laboratory of Industrial Control Technology, College of Control Science and Engineering, Zhejiang University, Hangzhou 310027, China, and Huzhou Institute, Zhejiang University, Huzhou 313000, China.}
}
\begin{document}

\maketitle

\begin{abstract}

This paper proposes Elastic Tracker, a flexible trajectory planning framework that can deal with challenging tracking tasks with guaranteed safety and visibility. Firstly, an object detection and intension-free motion prediction method is designed. Then an occlusion-aware path finding method is proposed to provide a proper topology. A smart safe flight corridor generation strategy is designed with the guiding path. An analytical occlusion cost is evaluated. Finally, an effective trajectory optimization approach enables to generate a spatio-temporal optimal trajectory within the resultant flight corridor. Particular formulations are designed to guarantee both safety and visibility, with all the above requirements optimized jointly. The experimental results show that our method works more robustly but with less computation than the existing methods, even in some challenging tracking tasks.

\end{abstract}
\section{Introduction}
\label{sec:introduction}

In recent years, with the advancement of vision and navigation technology, flying robots have been widely used in more and more complex missions. Autonomous aerial tracking is a challenging one appied in videography, chasing, cinematographer and surveillance.
Generally, there are three main technical challenges of aerial tracking: 
\begin{enumerate}
	\item[a)] Safety\label{elastic:safety}: The trajectory of the drone should be collision-free for both static obstacles and the target.
	\item[b)] Visibility\label{elastic:visibility}: The drone is supposed to keep the target in its limited FOV and avoid the target being occluded by surrounding obstacles.
	\item[c)] Smoothness: The trajectory should be smooth to avoid motion blur of the target in view.
\end{enumerate}

Some state-of-the-art aerial tracking planners \cite{jeon2020integrated, han2020fast, wang2021visibility} addressing the above issues have shown significant robustness and impressive agility. 
However, these methods are not sufficiently flexible to handle some extreme situations. For instance, when the target moves towards the drone suddenly, changes speed abruptly, and goes around the obstacles frequently, these methods are prone to fail.
Summarizing the reasons for the failure, most planners typically design a preplanning procedure like graph-searching to cover safety and visiblity, but carry out some refining work like path smoothing afterwards. Such inconsistency makes the obtained trajectories may not meet all the constraints. Besides, in some cases when the collision avoidance or dynamic feasibility contradicts the requirements of visibility, existing methods rarely obtain a feasible solution since the formulations of the constraints are not adaptive enough.
To achieve such tasks, an ideal tracking planner shall automatically trade-off the above requirements, which we call \emph{elasticity}. It is just like there's an invisible spring between the drone and the target, making them neither separated but also stretchable flexibly according to the situation.

In this paper, we propose \emph{Elastic Tracker}\footnote{\url{https://github.com/ZJU-FAST-Lab/Elastic-Tracker}}, a flexible tracking framework satisfying the mentioned requisites.
\begin{figure}[t]
    \begin{center}
        \includegraphics[width=1.0\columnwidth]{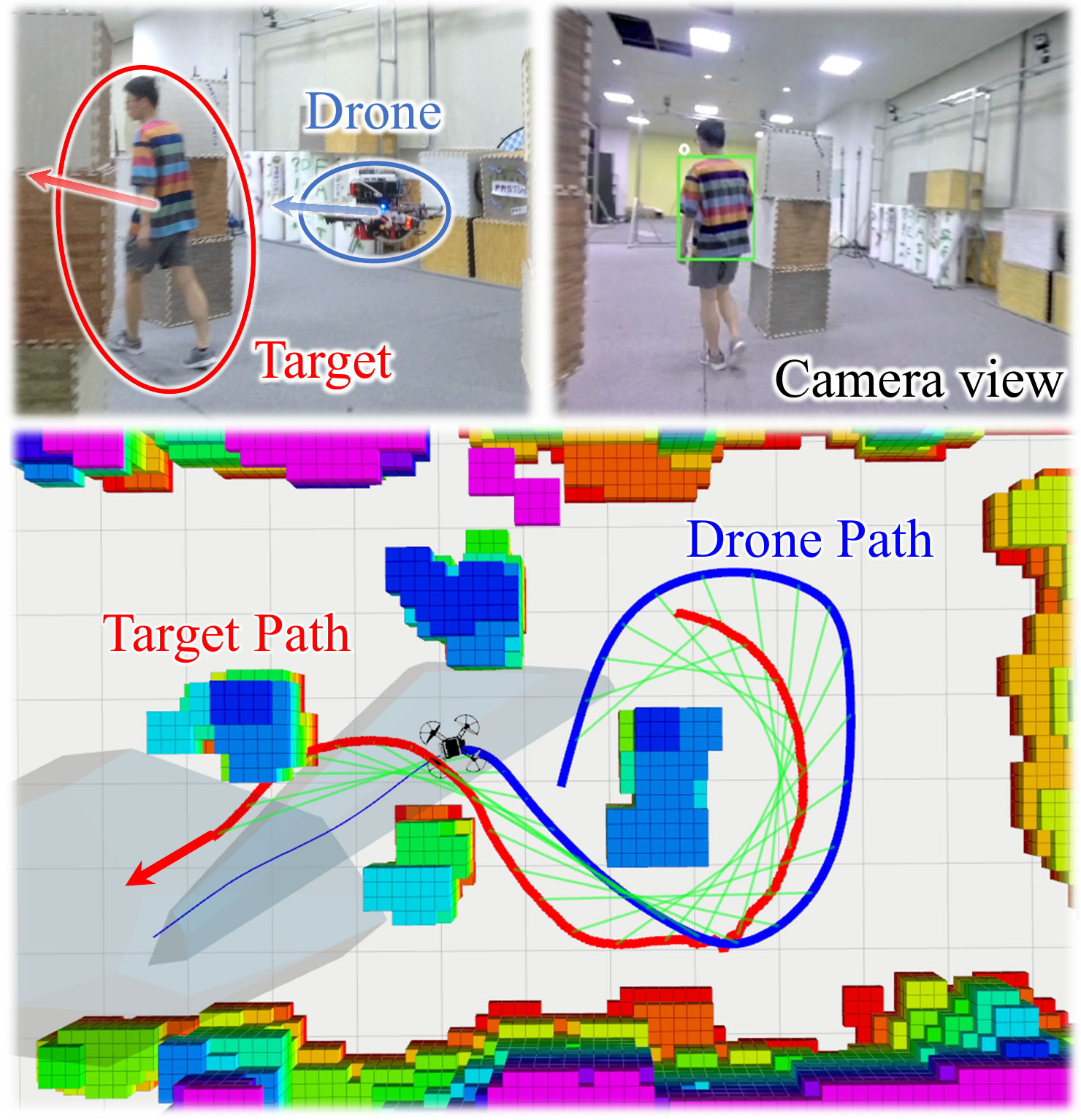}
    \end{center}
    \caption{
        \label{fig:introduction}
        Illustration of the performance of our Elastic Tracker in a real world experiment. The drone is able to keep a proper distance and avoid occlusion while the target moves around the obstacles.
    }
\vspace{-1cm}
\end{figure}
To begin with, we inherit the human detection and localization methods of our previous work\cite{pan2021fast}. Afterwards a lightweight intension-free motion prediction method is designed. Subsequently, an occlusion-aware path finding method is proposed to provide an appropriate topology. With the path's guaidance, a smart safe flight corridor generation strategy is designed, which considers the initial velocity of the drone. An analytical occlusion cost is evaluated passingly, which differs from \cite{jeon2020integrated, wang2021visibility} appraising occlusion with Euclidean Signed Distance Fields (ESDF). Thus the computational burden is greatly reduced. Finally, an effective joint optimization approach enables to generate a spatio-temporal optimal trajectory within the resultant flight corridor.
To achieve more elasticity, we design particular formulations for avoiding occlusion and keeping proper viewing distance. The proposed trajectory optimization approach is able to constrain the trajectory at both relative and absolute time while optimizing the time allocation of each piece.

We summarize our contributions as follows:
\begin{enumerate}
    \item An occlusion-aware path searching method and a smart safe flight corridor (SFC) generation strategy.
    \item An analytical occlusion cost is evaluated without constructing an ESDF using the result of the proposed path searching method.
    \item Particular formulations are designed for avoiding occlusion and keeping proper viewing distance.
    \item An effective trajectory optimization approach enables to generate a spatio-temporal optimal trajectory with guaranteed safety and visibility.
\end{enumerate}

\section{Related Work}
\label{sec:related_work}

Some vision-based tracking controllers \cite{kim2013vision, kendall2014board, cheng2017autonomous} take the tracking error defined on image space as the feedback. These reactive methods can achieve real-time performance but are short-sighted to consider safety and occlusion constraints.
Many previous research \cite{nageli2017real, penin2018vision, bonatti2018autonomous} employ a receding horizon formulation.
N$\ddot{\mathbf a}$geli et al. \cite{nageli2017real} propose a real-time receding horizon planner that optimizes both robot trajectories and gimbal controls for visibility under occlusion. Similarly, Penin et al. \cite{penin2018vision} design a non-linear model predictive control (NMPC) problem and solve it by sequential quadratic programming. However, the assumption \cite{nageli2017real, penin2018vision} that obstacles are all regarded as ellipsoids limits the application scenarios to artificial or structural environments. 

Bonatti et al. \cite{bonatti2018autonomous} trade-off shot smoothness, occlusion, and cinematography guidelines in a principled manner, even under noisy actor predictions. Nevertheless, they rely on numerical optimization of the entire objectives involving complex terms such as integration of signed distance field over a manifold, which cannot guarantee satisfactory optimality.
Jeon et al. \cite{jeon2019online, jeon2020detection, jeon2020integrated} propose a graph-search-based path planner along with a corridor-based smooth planner. The former generates a series of viewpoints, and the subsequent smooth planner follows the viewpoints. However, the preplanning procedure involves highly time-consuming graph construction. Additionally, this method assumes that the global map of the environment and the target’s moving intent are both known. Therefore, it cannot be used with general unknown environments and targets.
Han et al. \cite{han2020fast} propose a safe tracking trajectory planner consisting of a target informed kinodynamic searching front-end and a spatial-temporal optimal trajectory planning back-end. However, the optimization formulation of the latter trades off minimizing energy and time, which is entirely inconsistent with the original tracking problem.

Some perception-aware planners are applied to the scenarios resembling aerial tracking. Chen et al. \cite{chen2020bio} utilize a stereo camera with an independent rotational DOF to sense the obstacles actively. In particular, the sensing direction is planned heuristically by multiple objectives, including tracking dynamic obstacles, observing the heading direction, and exploring the previously unseen area. Watterson et al. \cite{watterson2020trajectory} plan pose and view direction of the camera by optimizing on manifold $ \mathbb R^3 \times S^2$. 
Furthermore, there are also some researches focusing on the planning of UAVs with a limited FOV.
Spasojevic et al. \cite{spasojevic2020perception} propose an optimal path reparameterization maintaining a given set of landmarks within FOV. PANTHER \cite{tordesillas2021panther} plans trajectories that avoid dynamic obstacles while also keeping them in FOV. However, all the methods above cannot handle the occlusion of static obstacles, which is rather critical for the perception of the target.

Wang et al. \cite{wang2021visibility} design an analytical visibility metric considering both distance keeping and occlusion. However, this metric relies on constructing an ESDF frequently, which is time-consuming. Besides, these methods construct such hard visibility constraints that they are prone to fail in cluttered environments. Notably, they plan both the position and orientation of the drone simultaneously to maximize the object detection, but ignore the observation of other obstacles, which causes unsafety.

%\section{Target Detection and Prediction}
%\label{sec:target}

%\subsection{Target Detection and Point Cloud Filter}
%
%For human detection, we apply open pose []. Then we track the target with an EKF under the assumption of the constant velocity.
%In order to avoid the target being regarded as a static obstacle while mapping, the surrounding point clouds of the target are removed manually. 

%\subsection{Target Motion Prediction}
%
%It's difficult to predict the moving target precisely without knowing its intention. Since our planner does not rely on accurate predictions, we just generate motion primitives for the target and choose a safe and minimum acceleration one. The future positions of the target are predicted as 
%\begin{align}
%  \label{eq:time_stamps}
%  \mathcal{T} = \cBrac{t_k \in [0, T_p] ~\Big |~ t_k \rightarrow z_k , 0 < k \leq M_\mathcal{T} }
%\end{align}

\section{Front-end Processing}

\subsection{Hierarchical Multi-goal Path Finding}
\label{sec:path}
\begin{figure}[ht]
	\begin{center} 
		\includegraphics[width=0.9\columnwidth]{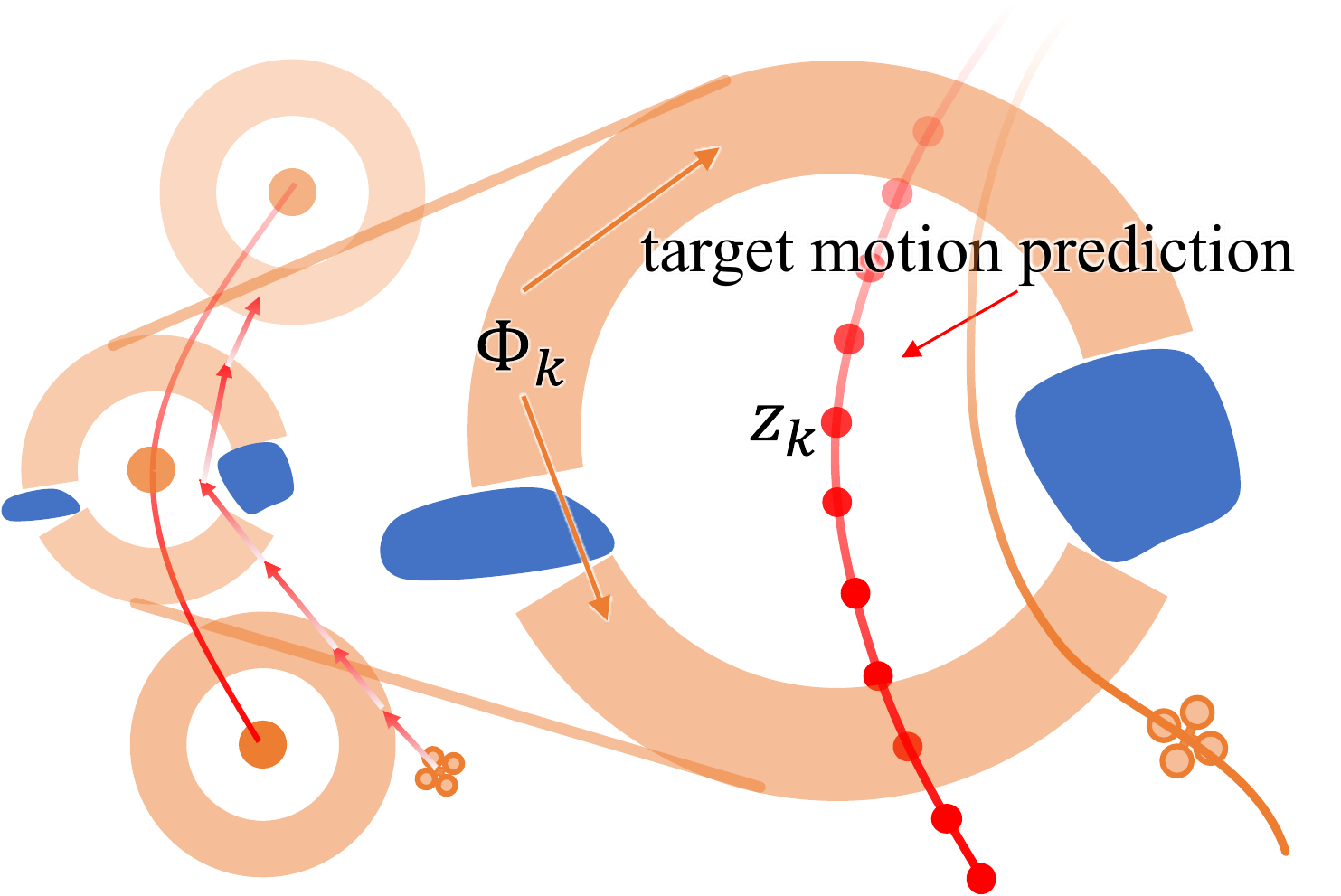}
	\end{center}
	\caption{
		\label{fig:astar}
		Given a series of future waypoints $z_k$ of the target, a path passing through the occlusion-free area $\Phi_k$ of each waypoint is found.
	}
\vspace{-0.5cm}
\end{figure}

Given a series of target future positions denoted by
\begin{align}
 \label{eq:time_stamps}
 \mathcal{T} = \cBrac{t_k \in [0, T_p] ~\Big |~ t_k \rightarrow z_k , 0 < k \leq M_\mathcal{T} }
\end{align}
Taking into account both distance and occlusion, we define an occlusion-free region $\Phi_k$ for each predicted position of the target $z_k$, shown in Fig. \ref{fig:astar}. In order to obtain a proper topology for subsequent actions, we aim to find a path passing through $\Phi_1, \Phi_2, ...$. Considering efficiency, we use the greedy method, decoupling it into smaller multi-goal path searching problems. $\Phi_k$ is set to the $k$-th $\mathbf A^\star$ destination region, according to which both cost function $f^k(n)$ and heuristic function $h^k(n)$ are designed as
\begin{subequations}
\begin{align}
  f^k(n) =& g^k(n) + h^k(n), \\
  h^k (n) =& \sqrt{\{d^k_{xy} (n)-d_d\}^2 + \{d^k_{z} (n)\}^2},
\end{align}
\end{subequations}
where $d^k_{xy}$ and $d^k_{z}$ are the horizontal and vertical components of the distance between node $n$ and $z_k$, $d_d$ is the desired distance between the drone and the target. Except for the colliding ones, the nodes within proper distance but occluded are set invalid. The stop condition is set as $n\in\Phi_k$. Each searching destination point $s_k$ is set to the next starting point.

\subsection{Safe Flight Corridor Generation}

Liu et al. \cite{liu2017planning} propose an efficient method for generating a safe flight corridor with a polygonal path. However, this method cannot handle infeasible initial states, and the generated polytopes would overlap too much with each other if the segments are too short.
We adopt this module and design an intelligent strategy to address the limitations. 

The first polytope is generated with a small segment from the initial position and along the direction of the initial speed. From the end of the first line segment, we search a guiding path as described in \ref{sec:path}. The next polytope is generated with the segment from the last end to the intersection of the last polytope and the guiding path. Finally, we can get a series of polytopes representing the safe region
\begin{align}
  \mathcal P = \bigcup_{i=1}^{M_\mathcal{P}}\mathcal{P}_i,
  \label{eq:HPolytopeDescription}
  ~\mathcal{P}_i =\cBrac{x\in\mathbb{R}^3~\Big|~\mathbf{A}_ix \leq b_i}.
\end{align}

\subsection{Visible Region Generation}

\begin{figure}[ht]
  \begin{center}
    \includegraphics[width=0.7\columnwidth]{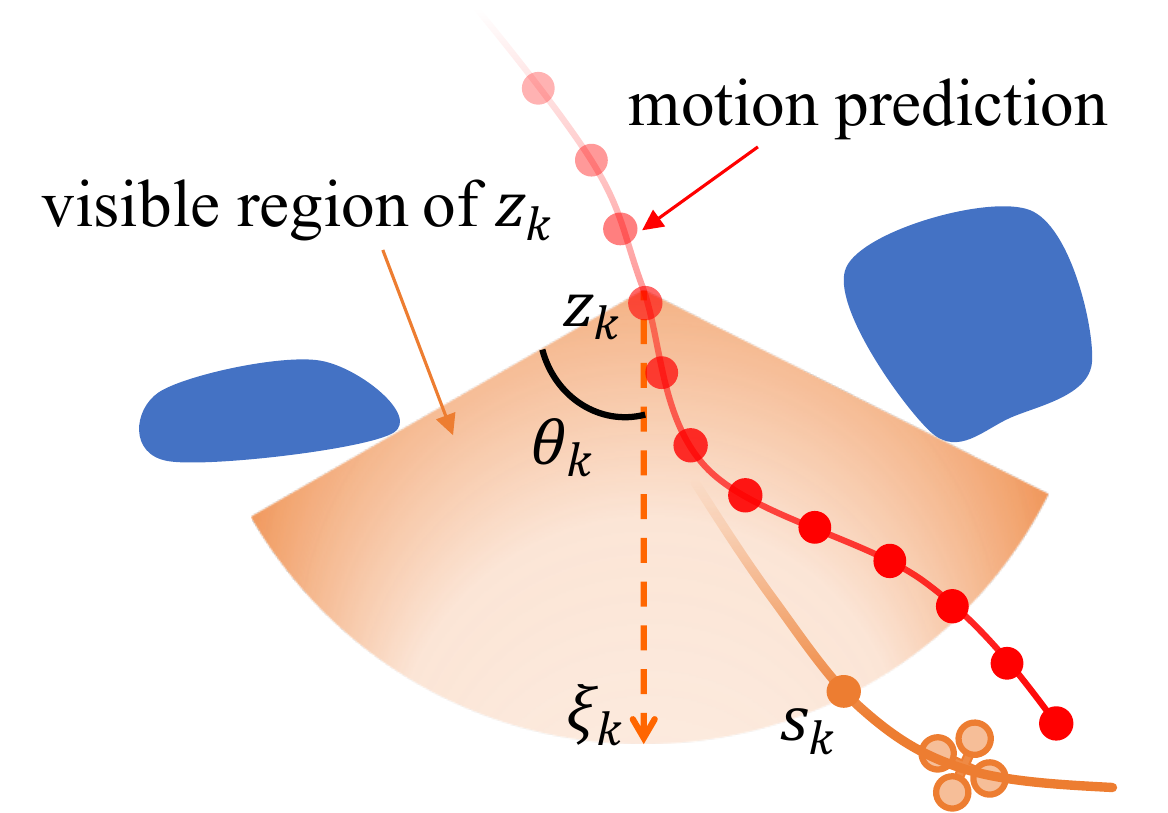}
  \end{center}
  \caption{
    \label{fig:visible_region}
    Visible region is represented as a sector-shaped area.
  }
\vspace{-0.5cm}
\end{figure}

For each target future position $z_k$ and a visible point $s_k$ as seed, we generate a sector-shaped visible region
\begin{align}
  \label{eq:visible_constraint}
  \mathcal V_k = & \cBrac{x\in \mathbb R^3~\Big|~\left\langle x-z_k, \xi_k \right\rangle \leq \theta_k },
\end{align}
where $\xi_k $ denotes the angle bisector, shown in Fig. \ref{fig:visible_region}.

\section{Trajectory Optimization}
\label{sec:optimization}

\subsection{MINCO Trajectory Class}

In this paper, we adopt $\mathfrak{T}_{\mathrm{MINCO}}$ \cite{wang2021geometrically}, a minimum control effort polynomial trajectory class defined as
\begin{align*}
  \mathfrak{T}_{\mathrm{MINCO}} = \Big\{ & p(t):[0, T]\mapsto\mathbb{R}^m~ \Big|~\mathbf{c}=\mathcal{M}(\mathbf{q},\mathbf{T}),~ \\
                                         & ~~\mathbf{q}\in\mathbb{R}^{m(M-1)},~\mathbf{T}\in\mathbb{R}_{>0}^M\Big\},
\end{align*}
where an $m$-dimensional trajectory $p(t)$ is represented by a piece-wise polynomial of $M$ pieces and $N=2s-1$ degree. The $i$-th piece is denoted by
\begin{equation}
  \label{eq:SinglePiece}
  p_i(t)=\mathbf{c}_i\tp\beta(t),~t\in[0,T_i],
\end{equation}
where $\mathbf{c}=\rbrac{\mathbf{c}_1\tp, \dots, \mathbf{c}_M\tp}\tp \in\mathbb{R}^{2Ms\times m}$, $\mathbf c_i \in \mathbb{R}^{2s\times m}$ is the coefficient matrix of the piece and $\beta(t)=\rbrac{1,t,\dots,t^N}\tp$ is the natural basis. Time vector $\mathbf{T}=\rbrac{T_1, \dots, T_M}\tp$, $T_i$ is the duration for the $i$-th piece.

All trajectories in $\mathfrak{T}_{\mathrm{MINCO}}$ have compact parameterization by only $\mathbf{q}$ and $\mathbf{T}$, where $\mathbf{q}=\rbrac{q_1,\dots,q_{M-1}}$, $q_i$ is the intermediate waypoint.  Evaluating an entire trajectory from $\mathbf{q}$ and $\mathbf{T}$ can be done via such a linear-complexity formulation
\begin{align}
  \mathbf c = \mathcal M(\mathbf q, \mathbf T),
  \label{eq:mc_b}
\end{align}
which allows any second-order continuous cost function $\mathcal F(\mathbf c, \mathbf T)$ with available gradient applicable to MINCO trajectories represented by $\mathbf q\& \mathbf T$. More specifically, the corresponding cost function for $\mathfrak{T}_{\mathrm{MINCO}}$ is computed as
\begin{align}
  \mathcal J(\mathbf q, \mathbf T) = \mathcal F(\mathcal M(\mathbf q, \mathbf T), \mathbf T)
\end{align}
Then the mapping Equ. \ref{eq:mc_b} gives a linear-complexity way to compute $\partial \mathcal J / \partial \mathbf q$ and $\partial \mathcal J / \partial \mathbf T$ from corresponding $\partial \mathcal F / \partial \mathbf c$ and $\partial \mathcal F / \partial \mathbf c$. After that, a high-level optimizer is able to optimize the objective efficiently.

% Besides, it should be noted that the trajectory obtained from mapping Equ. \ref{eq:mc_b} satisfies the following control effort minimization
% \begin{subequations}
%   \begin{align}
%     \min_{p(t)} & ~{\int_{t_0}^{t_M}{p^{(s)}(t)\tp \mathbf{W}p^{(s)}(t)}\df{t}}, \\
%     \mathit{s.t.}~   \label{eq:InitialTerminalConditions}
%                 & ~p^{[s-1]}(t_0)=\bar{p}_o,~p^{[s-1]}(t_M)=\bar{p}_f,           \\
%     \label{eq:IntermediateConditions}
%                 & ~p(t_i)=\bar{p}_i,~1\leq i<M,                                  \\
%                 & ~t_{i-1}<t_i,~1\leq i\leq M,
%   \end{align}
% \end{subequations}
% where $\mathbf{W}\in\mathbb{R}^{m\times m}$ is a diagonal matrix with positive entries, $\bar{p}_o\in\mathbb{R}^{ms}$ and $\bar{p}_f\in\mathbb{R}^{ms}$ are the initial and final condition, $\bar{p}_i \in\mathbb R^{m}$ is the given intermediate waypoint at time $t_i$.

%All the features of $\mathfrak{T}_{\mathrm{MINCO}}$ mentioned above are fully utilized in our senario.

\subsection{Problem Formulation}

We expect $T$, the total duration of the trajectory to be equal to $T_p$, the prediction duration of target. However, enforcing the drone to reach the final state in a fixed duration may cause dynamic infeasible in some cases, for instance, when the target moves faster than the chaser. Therefore, we make a time slack $T \geq T_p$ and set the objective as a tradeoff of minimum jerk and minimum time. The general tracking problem then is formulated as
\begin{subequations}
  \begin{align}
    \min_{p(t), T} & ~\mathcal J_o = \int_{0}^{T} {\norm{p^{(3)}(t)}^2} \df{t} + \rho T ,                                     \\
    s.t.~          & ~p^{[s-1]}(0)=\bar{p}_o,~p^{[s-1]}(T)=\bar{p}_f,                                                         \\
                   & ~\label{eq:dynamic_feasibility} \norm{p^{(1)}(t)}\leq v_m,~\norm{p^{(2)}(t)}\leq a_m,~\forall t\in[0,T], \\
                   & ~\label{eq:corridor_constraint} p(t)\in\mathcal{P},~\forall t\in[0,T],                                   \\
                   & ~\label{eq:visibility_constraint} p(t_k)\in\mathcal{V}_k,~\forall t_k \in \mathcal T,                    \\
                   & ~\label{eq:distance_constraint} d_l \leq \norm{p(t_k) - z_k} \leq d_u, \forall t_k \in \mathcal T,       \\
                   & ~\label{eq:time_slack} T \geq T_p,
  \end{align}
\end{subequations}
where $\rho$ is tunable parameter, $v_m$ and $a_m$ are velocity and acceleration bounds. Prediction timestamps $\mathcal T$, safe area $\mathcal P$ and visible region $\mathcal V_k$ are denoted in Equ. \ref{eq:time_stamps}, \ref{eq:HPolytopeDescription} and \ref{eq:visible_constraint}.

We use $\mathfrak{T}_{\mathrm{MINCO}}$ of $s=3$ for minimum jerk and $M = 2M_\mathcal P$ (2 pieces in each polytope) for enough freedom. Then the gradients $\partial{\mathcal J_o}/\partial \mathbf c$ and $\partial{\mathcal J_o}/\partial \mathbf T$ can be evaluated as
\begin{subequations}
  \begin{align}
    \frac{\partial{\mathcal J_o}}{\partial c_i} = & 2\left( \int_0^{T_i} \beta^{(3)}(t)\beta^{(3)}(t)\tp \df{t} \right) c_i, \\
    \frac{\partial{\mathcal J_o}}{\partial T_i} = & c_i\tp\beta^{(3)}(T_i)\beta^{(3)}(T_i)\tp c_i + \rho.
  \end{align}
\end{subequations}

Constraints Equ. \ref{eq:dynamic_feasibility} for dynamic feasbility and Equ. \ref{eq:corridor_constraint} for collision-free are handled with relative time integral penalty method, which is introduced in Section \ref{sec:relative_time}.
Constraints Equ. \ref{eq:visibility_constraint} for occlusion-free and Equ. \ref{eq:distance_constraint} for distance keeping are handled with absolute time penalty method, which is introduced in Section \ref{sec:absolute_time}. Constraint Equ. \ref{eq:time_slack} is eliminated with a transformation, introduced in Section \ref{sec:temporal_constraints_elimination}.

\subsection{Relative Time Integral Penalty Method}
\label{sec:relative_time}

Dynamic feasibility Equ. \ref{eq:dynamic_feasibility} and collision avoidance Equ. \ref{eq:corridor_constraint} constraints can be formulated with penalty function as follows
\begin{equation}
  \label{eq:SimpleContinuousConstraints}
  \begin{cases}
    \mathcal G_h = \mathbf A_i p_i(t)-b_i \leq \mathbf 0 , & \forall t\in[0,T_i], \\
    \mathcal G_v = \norm{p_i^{(1)}(t)}^2 - v_{m}^2 \leq 0, & \forall t\in[0,T_i], \\
    \mathcal G_a = \norm{p_i^{(2)}(t)}^2 - a_{m}^2 \leq 0, & \forall t\in[0,T_i]. \\
  \end{cases}
\end{equation}
Inspired by the constraint transcription \cite{jennings1990computational} method, $\mathcal G_\star$ can be transformed into finite constraints via integral of constraint violation, which is furthered transformed into the penalized sampled function $\mathcal J_I^\star$.

The constraints in Equ. \ref{eq:SimpleContinuousConstraints} are either linear or quadratic constraints for a specific $t$ and $i$, thus the time integral penalty with gradient can be easily derived then applied:
\begin{subequations}
  \begin{align}
    \label{eq:PieceTimeIntegralPenalty}
    \mathcal I^\star_i=                                             & \frac{T_i}{\kappa_i}\sum_{j=0}^{\kappa_i}\bar{\omega}_j\max{\sbrac{\mathcal{G}_\star \rbrac{\mathbf{c}_i,T_i,\frac{j}{\kappa_i}},\mathbf{0}}^3},              \\
    \mathcal J_I^\star =                                            & \sum_{i=1}^{M} \mathcal I^\star_i, ~\star = \{h, v, a\},                                                                                                      \\
    \frac{\partial \mathcal J^\star_I}{\partial c_i} =              & \frac{\partial \mathcal I^\star_i}{\partial \mathcal G_\star} \frac{\partial \mathcal G_\star}{\partial c_i},
    ~\frac{\partial \mathcal J^\star_I}{\partial T_i} =              \frac{\mathcal I^\star_i}{T_i} + \frac{j}{\kappa_i} \frac{\partial \mathcal I^\star_i}{\partial \mathcal G_\star}\frac{\partial \mathcal G_\star}{\partial t}, \\
    \frac{\partial \mathcal I^\star_i}{\partial \mathcal G_\star} = & 3\frac{T_i}{\kappa_i} \sum_{j=0}^{\kappa_i}\bar{\omega}_j\chi\tp\max{\sbrac{\mathcal{G}_\star \rbrac{\mathbf{c}_i,T_i,\frac{j}{\kappa_i}},\mathbf{0}}^2},
  \end{align}
\end{subequations}
where $\rbrac{\bar{\omega}_0,\bar{\omega}_1,\dots,\bar{\omega}_{\kappa_i-1},\bar{\omega}_{\kappa_i}}=\rbrac{1/2,1,\cdots,1,1/2}$ are the quadrature coefficients following the trapezoidal rule~\cite{press2007numerical}.

% \begin{subequations}
%   \begin{align}
%     \frac{\partial \mathcal J^\star_I}{\partial c_i} =  \frac{\partial \mathcal I^\star_i}{\partial \mathcal G_\star} \frac{\partial \mathcal G_\star}{\partial c_i},
%     ~\frac{\partial \mathcal J^\star_I}{\partial T_i} =  \frac{\mathcal I^\star_i}{T_i} + \frac{j}{\kappa_i} \frac{\partial \mathcal I^\star_i}{\partial \mathcal G_\star}\frac{\partial \mathcal G_\star}{\partial t}, \\
%     \frac{\partial \mathcal I^\star_i}{\partial \mathcal G_\star} = 3\frac{T_i}{\kappa_i} \sum_{j=0}^{\kappa_i}\bar{\omega}_j\chi\tp\max{\sbrac{\mathcal{G}_\star \rbrac{\mathbf{c}_i,T_i,\frac{j}{\kappa_i}},\mathbf{0}}^2},
%   \end{align}
% \end{subequations}

\subsection{Absolute Time Penalty}
\label{sec:absolute_time}

Since the motion prediction of target is represented by a series of discrete waypoints and the visible region is also represented by a series of discrete circular sector, both constraints Equ. \ref{eq:visibility_constraint} and Equ. \ref{eq:distance_constraint} should be applied to the positions at some specified time $t_k \in \mathcal T$. For a position $p(t_k)$, assume that $t_k$ is on the $j$-th piece of the trajectory, i.e.
\begin{align}
   & ~\sum_{i=1}^{j-1} T_i \leq t_k \leq \sum_{i=1}^j T_i.
\end{align}

Then the gradients of $\mathbf c$ and $\mathbf T$ can be evaluated as
\begin{subequations}
  \begin{align}
    \frac{\partial p}{\partial c_i} = &
    \begin{cases}
      \beta(t_k - \sum_{i=1}^{j-1} T_i), ~ & i = j,    \\
      0, ~                                 & i \neq j,
    \end{cases}          \\
    \frac{\partial p}{\partial T_i} = &
    \begin{cases}
      -\dot{p}(t_k), ~ & i < j,    \\
      0, ~             & i \geq j,
    \end{cases}
  \end{align}
\end{subequations}

However, the duration of each piece of the trajectory $T_i$ is to be optimized and which piece $j$ that $t_k$ belongs to will change as optimization processes. Thus the cost and gradient will be discontinuous, which makes the optimization fail.

Fortunately, although the costs and gradients of $\mathbf c$ and $\mathbf T$ will be discontinuous, the costs and gradients of $\mathbf q$ and $\mathbf T$ are continuous using Equ. \ref{eq:mc_b}. The detailed proof is omitted due to the limited space.

\subsubsection{Distance-keeping Constraints} 
\begin{figure}[t]
  \begin{center}
    \includegraphics[width=0.8\columnwidth]{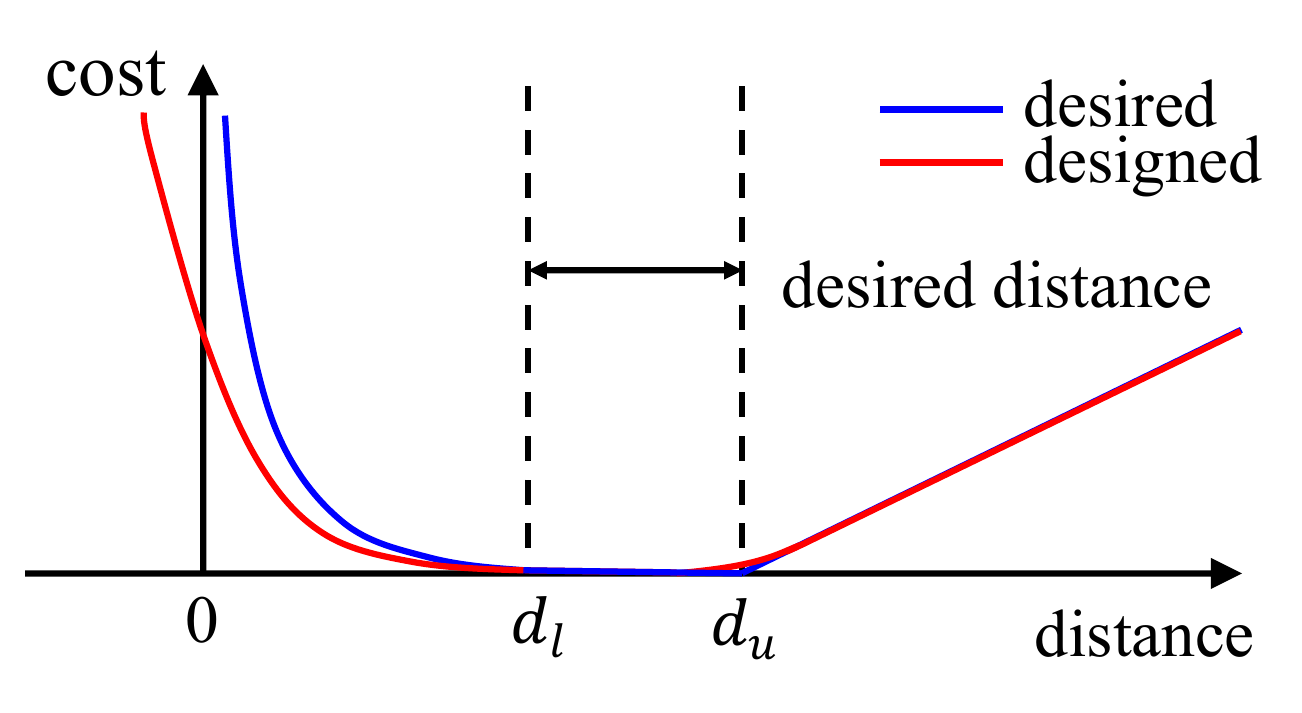}
  \end{center}
  \caption{
    \label{fig:keep_distance}
    To avoid contradicting the safety constraint, we design such a cost function for horizontal distance between the drone and the target, which rises abruptly when it's too small but increases linearly when it's too large.
  }
  \vspace{-1cm}
\end{figure}

Considering both safety and visibility, it's necessary for the drone to keep a proper distance from the target.
Since the drone lacks the independent DOF of pitch, we design different costs for vertical ($\delta^v$) and horizontal ($\delta^h$) components of the distance. For the former we set a small vertical tolerance such that $\delta^v \leq \delta^v_d$. For the latter, we design a $C^2$ penalty function $\mathcal H_d(x) =$
\begin{equation}
  \label{eq:keepDistance}
  \begin{cases}
    (x-d_l)^3,                                               & x \leq d_l,                        \\
    0,                                                       & d_l \leq x < d_u-\epsilon,         \\
    \frac{1}{\epsilon^3}(x-d_u+\epsilon)^3(3\epsilon-x+d_u), & -\epsilon \leq x - d_u < \epsilon, \\
    16(x-d_u),                                               & x > d_u+\epsilon,
  \end{cases}
\end{equation}
shown in Fig. \ref{fig:keep_distance}, where $[d_l, d_u]$ are range of desired distance and $\epsilon$ is a tiny constant.

Thus the cost of this term can be writen as
\begin{align}
  \mathcal J_a^d = \sum_{k=1}^{M_\mathcal T}~ \mathcal H_d \left(\delta^h_k\right) + \max \{\delta^v_k- \delta^v_d, 0\}^3,
\end{align}
where $\delta^h_k$ and $\delta^v_k$ are respectively the horizontal and vertical distance between $p(t_k)$ and $z_k$.

\subsubsection{Occlusion-free Constraints}
Given a tunable angle clearance $\theta_\epsilon$, visible region constraint Equ. \ref{eq:visible_constraint} can be writen as
\begin{align}
  \mathcal G_o(x, k) = \cos(\theta_k-\theta_\epsilon) - \frac{\overrightarrow{x-z_k} \cdot \overrightarrow{\xi_k}}{\norm{x-z_k}} \leq 0,
\end{align}
Apply the same penalty function in Equ. \ref{eq:SimpleContinuousConstraints} to this term
\begin{align}
  \mathcal J_a^o = \sum_{k=1}^{M_\mathcal T} ~\max~[\mathcal G_o(p(t_k), k), 0]^3.
\end{align}

Denote $\mathbf p = (p(t_1), ..., p(t_{M_\mathcal T}))$, then the gradients of both $\mathcal J_a^d$ and $\mathcal J_a^v$ can be calculated with $\mathcal O(M_\mathcal T)$ complexity by
\begin{align}
  \frac{\partial \mathcal J_a^\star}{\partial c_i} = \frac{\partial \mathcal J_a^\star}{\partial \mathbf p} \frac{\partial \mathbf p}{\partial c_i},
  ~\frac{\partial \mathcal J_a^\star}{\partial T_i} = \frac{\partial \mathcal J_a^\star}{\partial \mathbf p} \frac{\partial \mathbf p}{\partial T_i},
  ~\star = \{d,v\}.
\end{align}

\subsection{Temporal Constraints Elimination}
\label{sec:temporal_constraints_elimination}

Time slack constraint Equ. \ref{eq:time_slack} can be writen as
\begin{align}
  \sum_{i=1}^{M} T_i \geq T_p.
\end{align}

Inspired by \cite{wang2021geometrically}, we denote $\boldsymbol{\tau} = (\tau_1, ..., \tau_M) \in \mathbb{R}^M$ as new variables to be optimized and use the transformation
\begin{subequations}
  \begin{align}
    T_{\Sigma} & = T_p + \tau_M^2                                                           \\
    T_i        & =\frac{e^{\tau_i}}{1+\sum_{j=1}^{M-1}{e^{\tau_j}}}T_\Sigma, ~ 1 \leq i < M \\
    T_M        & =T_\Sigma-\sum_{i=1}^{M-1}{T_i}.
  \end{align}
\end{subequations}
Thus the temporal constraints are eliminated using such substitution.

% \section{Yaw Planning}

% Since the yaw is independent, we carry out yaw planning separately.
% We constrain the angular velocity of yaw and make the drone head in the direction towards the target. Besides, we check whether the trajectory for a period in the future enters unknown area and make the drone head in the direction towards the unknown area if true.

\section{Experiments and Benchmarks}
\label{sec:experiments}

\subsection{Implementation Details}

For human detection, we apply open pose \cite{cao2019openpose} and track the target with an EKF under the assumption of the constant velocity.
In order to avoid the target being regarded as a static obstacle while mapping, the surrounding point clouds of the target are removed manually. 
As for the motion prediction of the target, we just generate motion primitives for the target and choose a safe and minimum acceleration one.

Since the yaw is independent, we carry out yaw planning separately.
We constrain the angular velocity of yaw and make the drone head in the direction towards the target. Besides, we check whether the trajectory for a period in the future enters the unknown area and make the drone head in the direction towards the unknown area if true.

\subsection{Real World Experiments}

\begin{figure}[t]
  \begin{center}
    \includegraphics[width=1.0\columnwidth]{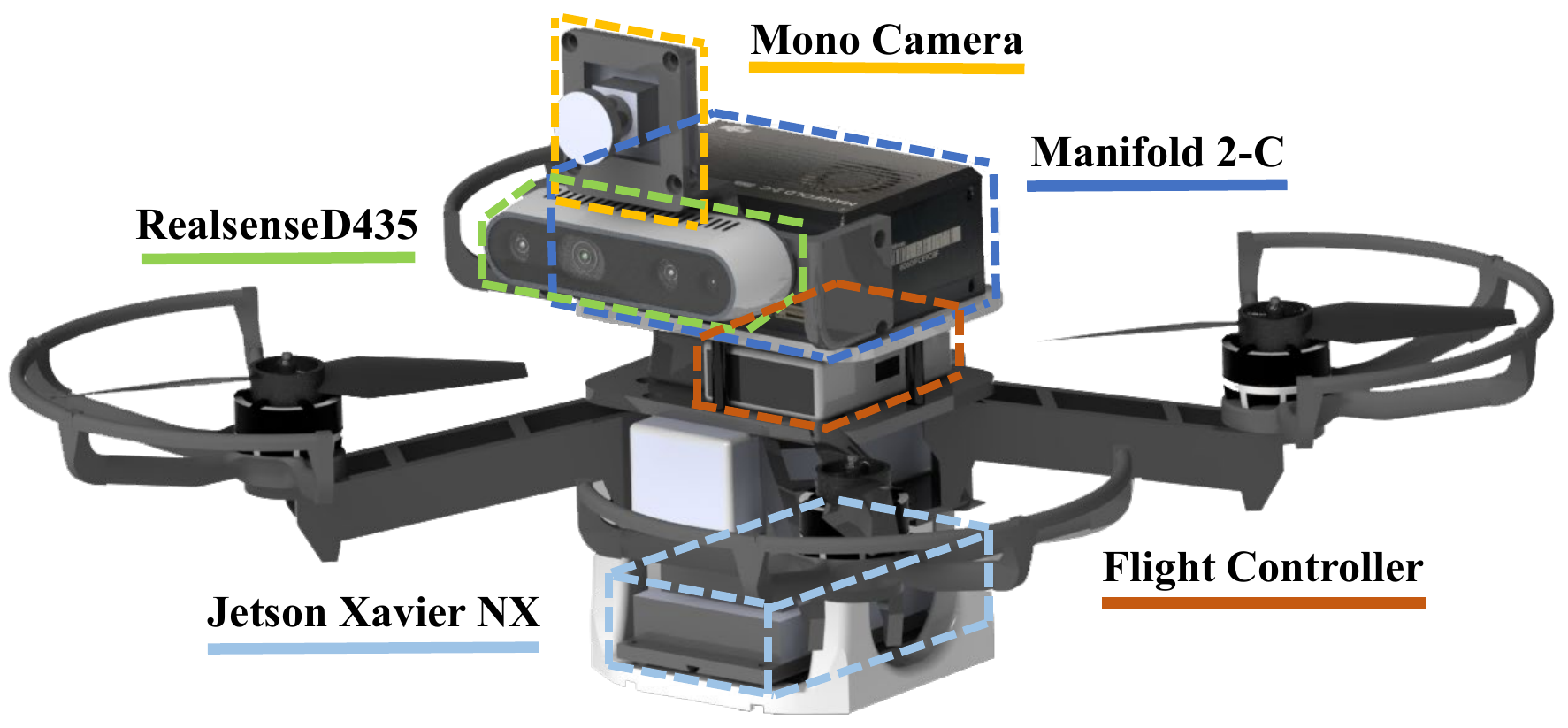}
  \end{center}
  \caption{
    \label{fig:hardware}
    Quadrotor platform used in our experiment.
  }
  \vspace{-1cm}
\end{figure}

Our quadrotor platform, shown in Fig. \ref{fig:hardware}, is equipped with an Intel RealSense D435 depth camera for self-localization \cite{qin2018vins} and mapping, a mono camera (FOV = $85^\circ \times 72^\circ$) for shooting the target, a Jetson Xavier NX4 for running object detection and a Manifold2 for the other computation tasks.

\begin{figure}[t]
  \begin{center}
    \includegraphics[width=0.9\columnwidth]{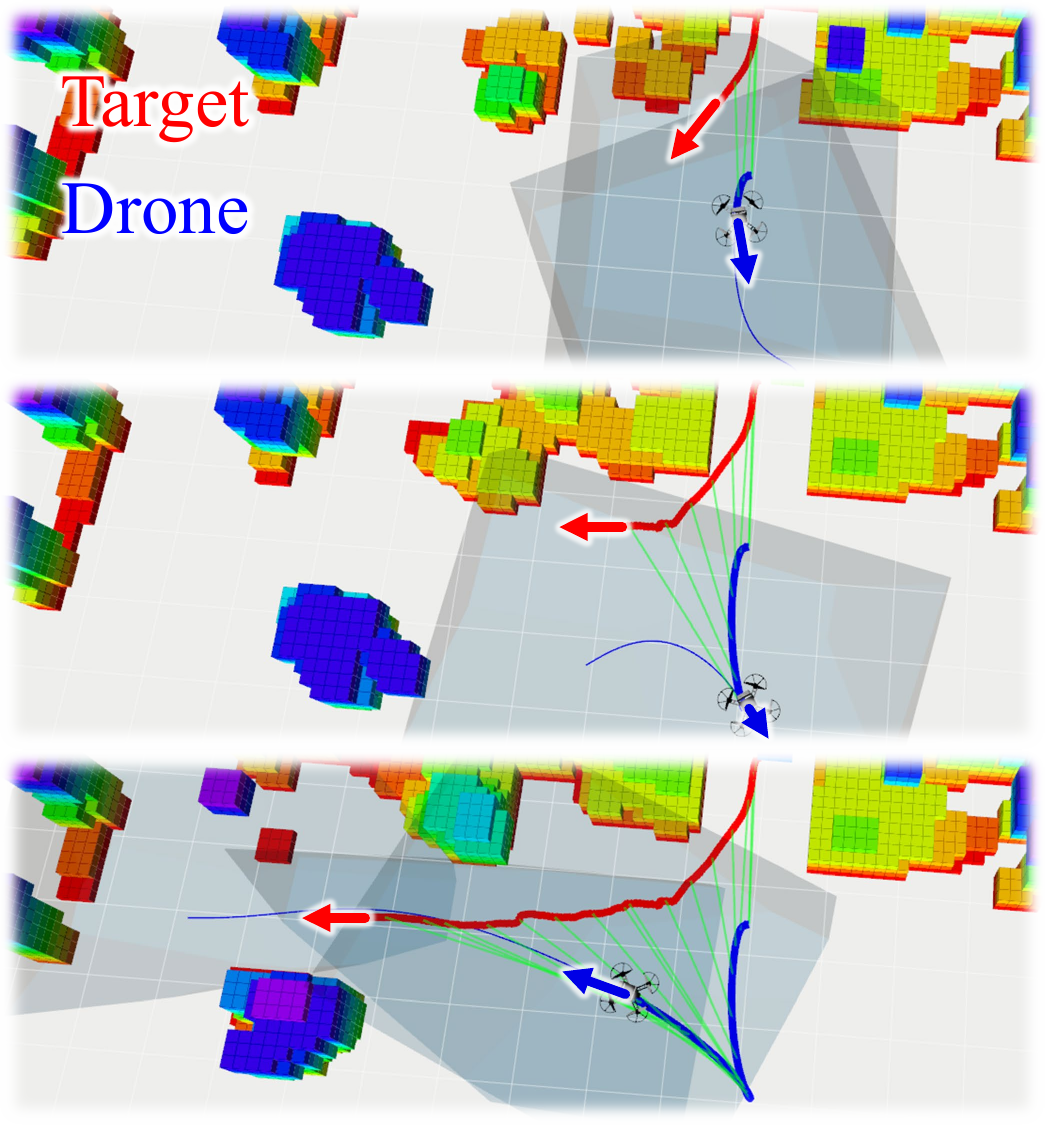}
  \end{center}
  \caption{
    \label{fig:exp_back}
    While the target moves close or away from the drone, a proper distance is kept, which is just like there's an invisible spring between them.
  }
\end{figure}

\begin{figure}[t]
  \begin{center}
    \includegraphics[width=0.9\columnwidth]{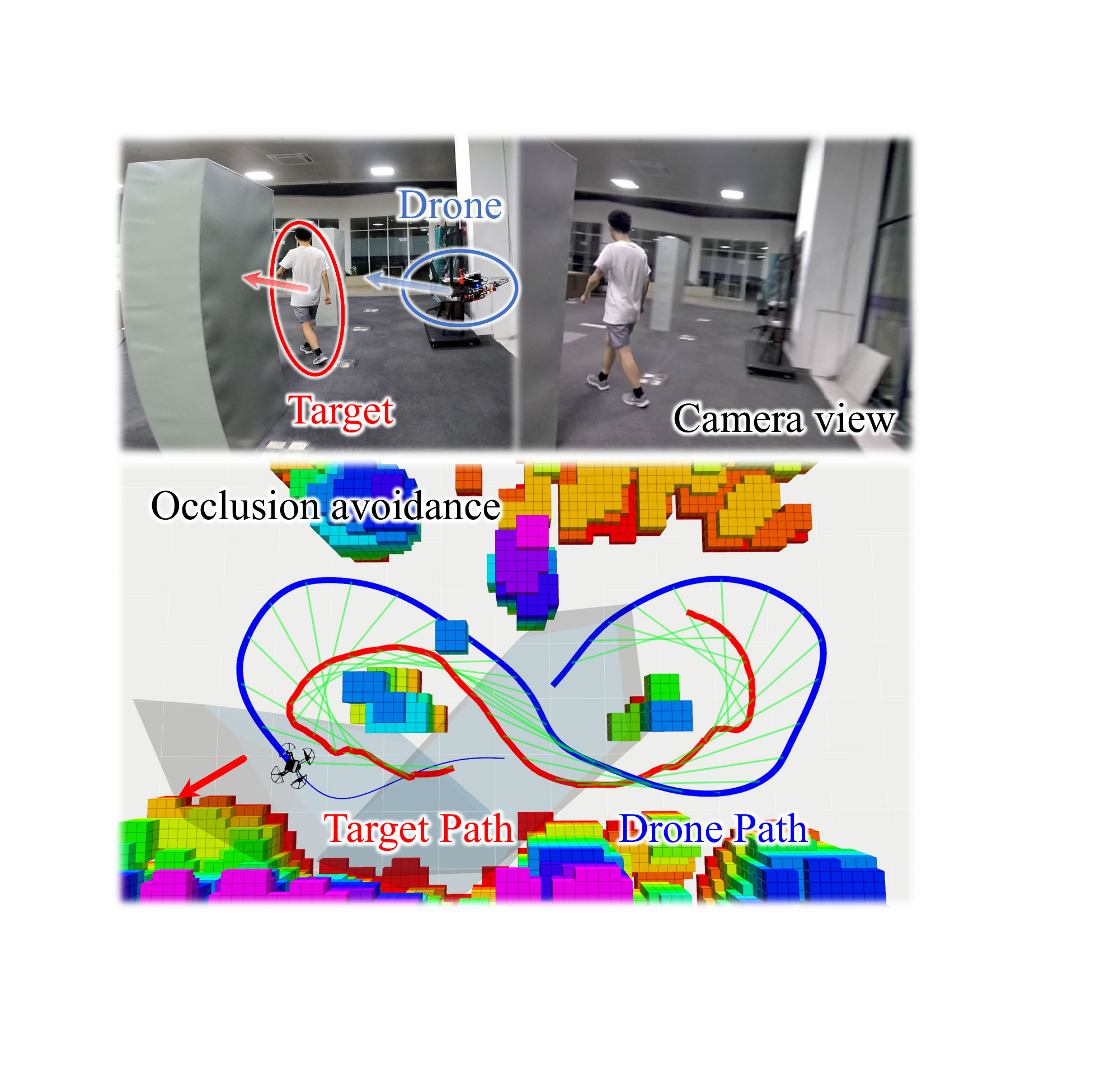}
  \end{center}
  \caption{
    \label{fig:exp_occlusion}
    While the target walks a figure eight around two obstacles, the drone can avoid occlusion and keep the target in the view.
  }
\vspace{-1.5cm}
\end{figure}

To validate the robustness of our method, we set up experiments in several challenging scenarios. In a long-term tracking experiment, the target moves towards the drone suddenly, shown in Fig. \ref{fig:exp_back} and the drone is able to keep a safe distance from the target flexibly. In a visibility test, the target walks a figure eight around two cabinets, shown in Fig. \ref{fig:exp_occlusion} and the drone is able to keep the target in the FOV.

\subsection{Simulation and Benchmark Comparisons}

\begin{figure*}[t]
	\begin{center}
		\subfigure[\label{fig:target_path}]{
			\includegraphics[height=0.58\columnwidth]{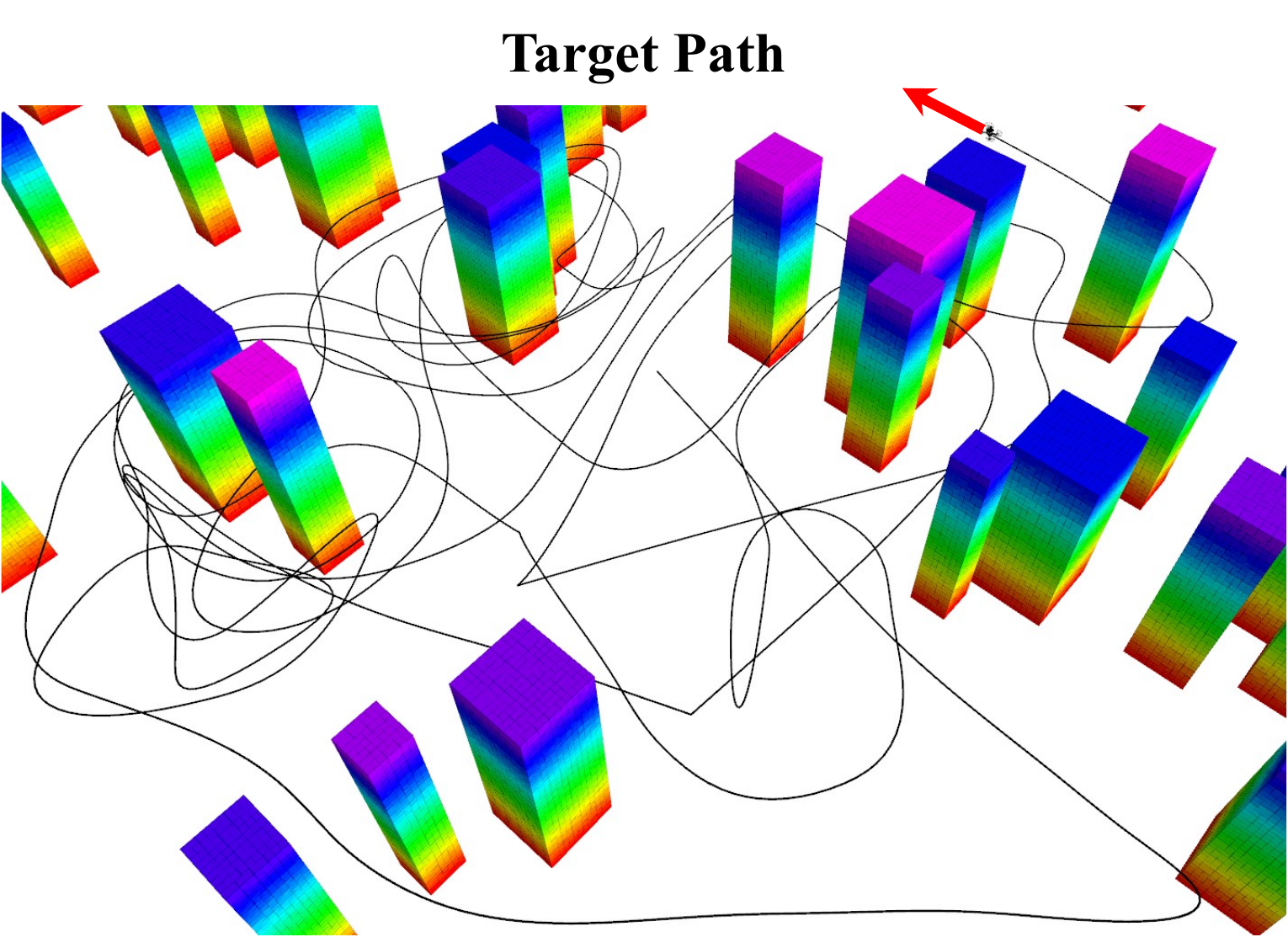}
		}
		\subfigure[\label{fig:projected_v}]{
			\includegraphics[height=0.58\columnwidth]{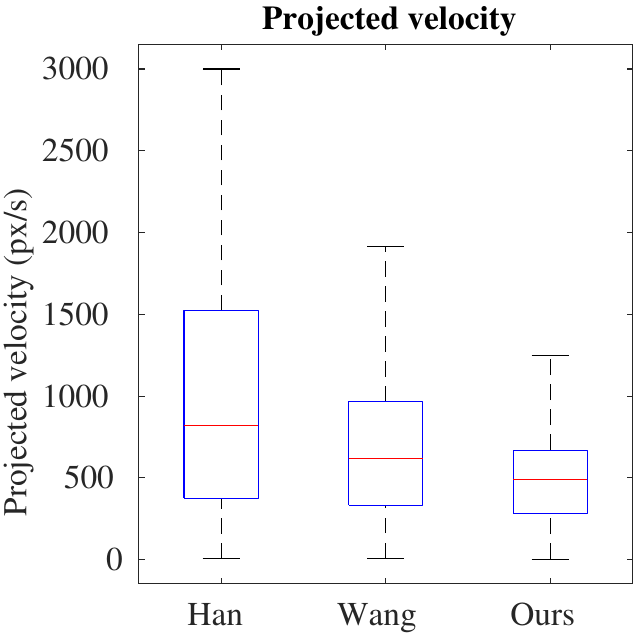}
		}
		\subfigure[\label{fig:failue_time}]{
			\includegraphics[height=0.58\columnwidth]{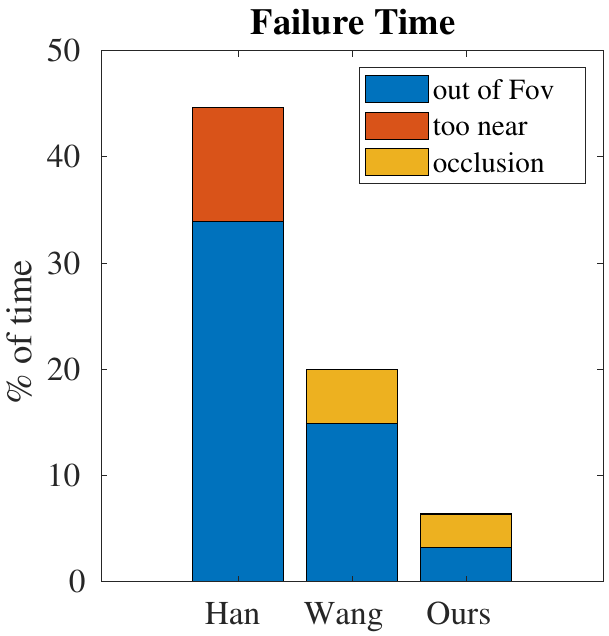}
		}
	\end{center}
	\caption{
		(a) History path of the target while moving aggressively in a cluttered environment. (b) The velocity of the projection of the centroid of the target onto the image plane. (c) Failure time of three different cases: out of view, too near (distance less than 1m), occluded by surrounding obstacles.
	}
\end{figure*}

\begin{figure*}[t]
	\begin{center}
		\includegraphics[height=0.65\columnwidth]{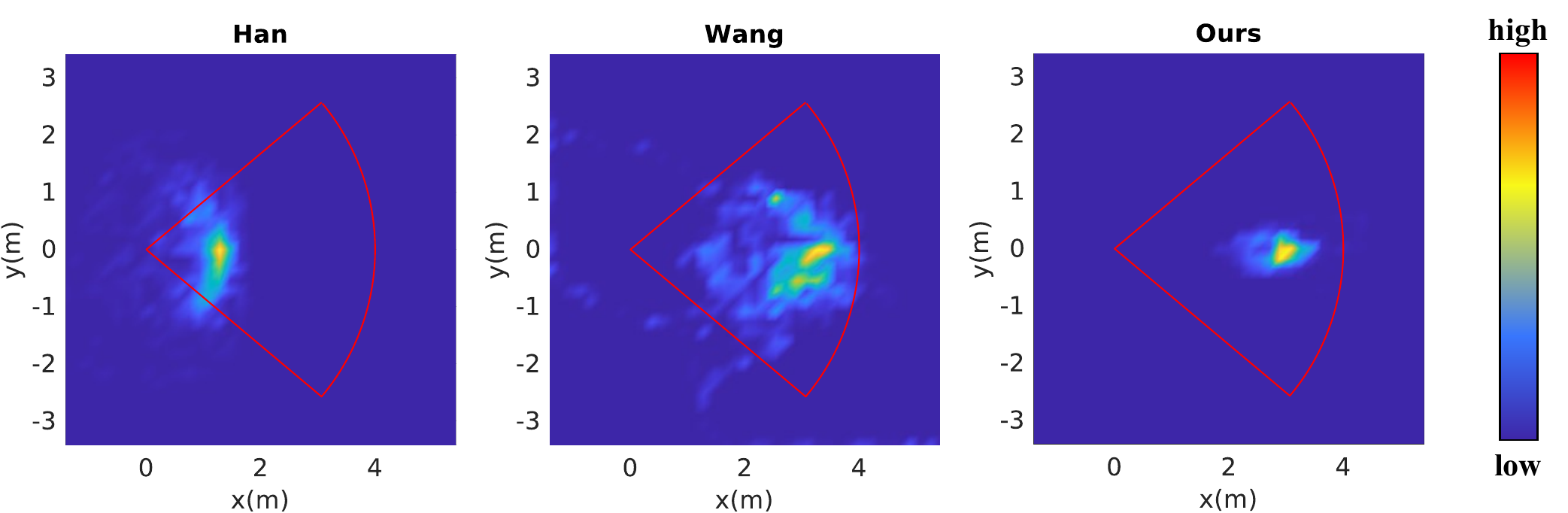}
	\end{center}
	\caption{
		\label{fig:heatmap}
		The distribution of the target positions relative to the tracking quadrotor on x-y plane. The red sector represents the FOV of the drone.
	}
\end{figure*}

We benchmark our method with Han's\cite{han2020fast} and Wang's \cite{wang2021visibility} work in simulation. All the simulation experiments are run on a desktop equipped with an Intel Core i7-6700 CPU. To compare the tracking performance fairly, we set three drones using each method to chase the same target simultaneously.  The target moves along such an aggressive path in a cluttered environment shown in Fig. \ref{fig:target_path}. The max velocity of the target is set as $2m/s$, and its position is broadcasted to each chaser. The max velocity and acceleration of the chasers are set as $3m/s$ and $6m/s^2$. The camera has an image size of 640×480 px$^2$, a limited FOV of $80^\circ \times 65^\circ$. 

The velocity of the projection of the target onto the image plane is shown in Fig. \ref{fig:projected_v}, which means our method achieves a much less blurred projection of the target than the other two methods.
We count the failure time of each chaser shown in Fig. \ref{fig:failue_time}. The failures can be categorized under three headings: out of FOV, too near from the drone, occluded from the surrounding obstacles. We can see that the failure rate of our method is much lower than the other two methods.
Furthermore, we count the target positions projected to x-y plane in the tracking quadrotor’s FOV, shown in Fig. \ref{fig:heatmap}. The heat map shows the distribution of the target positions relative to the drone, where our method is able to keep the obstacle inside the FOV limits much better.

We also benchmark the computation time of the three methods in different scenarios. As is shown in Tab. \ref{table:time}, the proposed method needs a much lower computation budget.

\begin{table}[ht]
	\renewcommand\arraystretch{1.1}
	\centering
	\caption{Computation time Comparison}
	\label{table:time}
	\begin{tabular}{c c c c c c c}
		\toprule
		{Target}                & \multirow{2}{*}{Method} & \multicolumn{5}{c}{Average calculating time (ms)}                                                                                  \\ \cmidrule(lr){3-7}
		$v_\text{max}$          &                         & $t_\text{path}$                                   & $t_\text{corridor}$ & $t_\text{ESDF}$  & $t_\text{optimze}$ & $t_\text{total}$ \\ \midrule
		\multirow{3}{*}{$1m/s$} & Han                     & 6.67                                              & 2.89                & \textbackslash{} & 0.38               & 9.94             \\
		& Wang                    & 0.63                                              & \textbackslash{}    & 6.43             & 5.5                & 12.56            \\
		& Ours                    & 0.856                                         & 0.538           & \textbackslash{} & 1.79           & \bf 3.184        \\ \midrule
		\multirow{3}{*}{$2m/s$} & Han                     & 10.5                                              & 4.62                & \textbackslash{} & 0.44               & 15.56            \\
		& Wang                    & 0.54                                              & \textbackslash{}    & 6.43             & 5.7                & 12.67            \\
		& Ours                    & 1.31                                          &  1.02            & \textbackslash{} & 2.87           & \bf 5.2          \\ \bottomrule
	\end{tabular}
\end{table}

\section{Conclusion and future work}
\label{sec:conclusion}

In this paper, we summarize the challenging requirements of elastic tracking and propose a flexible trajectory planning framework for aerial tracking. Extensive simulations and real-world experiments validate the robustness and efficiency of the proposed method. In the future, we will estimate the target's intention and improve the motion prediction. Furthermore, we will generalize our method to more extreme scenarios like tracking escaping target.

% \addtolength{\textheight}{-5.6cm} 
\bibliographystyle{IEEEtran}
\bibliography{ICRA2022jjl}
\end{document}